\documentstyle[12pt,fullpage]{article}
\input{psfig}

\def\cutwspace{\setlength{\parskip}{0pt}\setlength{\itemsep}{1pt}}

\title{\bf NetNeg: A Connectionist-Agent Integrated System for Representing
Musical Knowledge}
\author{Claudia V. Goldman \thanks{Supported by the Eshkol Fellowship,
Israeli Ministry of Science} \and Dan Gang\thanks{Supported by the
Eshkol Fellowship, Israeli Ministry of Science} \and Jeffrey S.~Rosenschein
\and Daniel Lehmann\\
       Insitute of Computer Science \\
       The Hebrew University\\
       Givat Ram, Jerusalem, Israel\\
       ph: 011-972-2-658-5353\ 
       fax: 011-972-2-658-5439\\
       email:clag@cs.huji.ac.il, dang@cs.huji.ac.il, jeff@cs.huji.ac.il,
lehmann@cs.huji.ac.il \\
       url: http://www.cs.huji.ac.il/\{$\sim clag,\sim dang,\sim jeff$\}
 }
\date{}
\begin{document}
\maketitle
\begin{abstract}

The system presented here shows the feasibility of modeling the
knowledge involved in a complex musical activity by integrating
sub-symbolic and symbolic processes.
This research focuses on the question of whether there is any advantage
in integrating a neural network together with a distributed artificial 
intelligence approach within the music domain.

The primary purpose of our work is to design a model that describes
the different aspects a user might be interested in considering when
involved in a musical activity. The approach we suggest in this work
enables the musician to encode his knowledge, intuitions, and
aesthetic taste into different modules. The system captures these
aspects by computing and applying three distinct functions: rules,
fuzzy concepts, and learning.

As a case study, we began experimenting with first species two-part
counterpoint melodies. We have developed a hybrid system composed of a
connectionist module and an agent-based module to combine the
sub-symbolic and symbolic levels to achieve this task.  The technique
presented here to represent musical knowledge constitutes a new
approach for composing polyphonic music.

\end{abstract}


\section{Introduction}

Researchers in computer music have chosen to use techniques from
Artificial Intelligence (AI) to explore complex musical tasks at the
cognitive level. These include tasks such as composition, listening,
analysis, and performance.  Research in AI can also benefit from
results in music research; in music, for example, aspects such as time
and hierarchical structure are inherent to the domain.

Simulating and modeling a musician's activities are tasks that are
appropriate for experimentation within the framework of artificial
intelligence.  The cognitive processes a musician undergoes are
complex and non-trivial to model.  Whenever we are involved in any
musical activity, we are faced with symbolic and sub-symbolic
processes. While listening, our aesthetic judgment cannot be applied
by following explicit rules solely.  Applying a learning mechanism to
model the task of listening~\cite{BT91,GB96,BG96,BG97} has been shown
to be a convenient (and appropriate) way to overcome the explicit
formulation of rules. Nevertheless, some of these aesthetic processes
might have a symbolic representation and might be specified by a
rule-based system~\cite{JAC91}.


The contribution of this work to AI is in the area of knowledge representation.
To demonstrate the advantages and computational power of our hybrid
knowledge representation, we design a model that describes the
different aspects a user might be interested in considering when involved
in a musical activity. The approach we suggest in this work enables
the musician to encode his knowledge, intuitions, and aesthetic taste
into different modules. The system captures these aspects by computing
and applying three distinct functions: rules, fuzzy concepts, and
learning.

Creating contrapuntal music in real time is a complex and challenging
domain of research. A single paradigm may not be sufficient to deal
with this kind of music. We therefore suggest using this music
creation task as a domain in which to test the feasibility of our hybrid
knowledge representation.

By choosing a specific task, we can examine the performance of the
hybrid system.  In addition to the theoretical interest in exploring
such an architecture, we believe that this composition system can
serve as an interactive tool for a composer, or as a real time
performance tool.

The specific case study we have chosen to experiment with is the polyphonic 
vocal style of the Sixteenth Century; more specifically, we investigate 
two-part species counterpoint (i.e, bicinia).

\paragraph{First Species Counterpoint:} this is the first species, the most
restrictive one, out of five species defined in species counterpoint
music.
Following~\cite{DictMusic}:
\begin{quote}
The progressive arrangement of the method, in dialogue form, with the rules
for each species dependent more or less on the restrictions of the preceding
was widely admired for its pedagogical value.
\end{quote}

In first species counterpoint, the cantus firmus (i.e., one of the 
two voices) is given in a whole note representation. The counterpoint
(i.e., the second voice) is created by matching a note against another note
in the cantus firmus. This matching follows specific rules as we explain
in Section~\ref{agents}. 

The specific task we examine is different in some important respects
from the cognitive task of composing counterpoint melodies. First, our
system creates {\em both\/} parts of the melody. Second, we do not
incorporate any backtracking process, since we deal with composition in
real time.
Nevertheless, the system architecture exploits the dynamic context to
produce the proper continuation of a melody (i.e., the already
chosen notes in the past influence the future choices in real time).

{\em NetNeg}, is composed of two main sub-systems. One sub-system is
implemented by a modified version of Jordan's sequential neural
network~\cite{Jordan86}. This net learns to produce new melody parts.
The second sub-system is a two-agent model based on Distributed
Artificial Intelligence~\cite{DAIr88}. These agents compose the two
parts of the melody according to the rules of the style.  These agents
negotiate with one another to maximize the global utility of the whole
system, which for our purposes should be interpreted as the global
quality of the composition.

In spite of the simplicity of the problem, our approach can serve as a
basis for further investigation of more complex musical problems.
This simple domain could have been formulated in a rule based system.
Nevertheless, our aim was to choose a starting domain that we could
evaluate and in which we could control the complexity of the process.
Our hope is that in more complex tasks, our system would be able to
integrate the performance of both symbolic and sub-symbolic musical
knowledge, as will be shown for the simpler case in the following
sections.

In this work, we want to emphasize the promising potential of using
the hybrid system approach for building models of musical activities
at the cognitive level.
Moreover, dealing with the problem of creating music in real time
opens up the possibility of building real time interactive
applications that combine the activities of a musician and a computer.

The paper is organized as follows. First, we present background information
regarding AI methodologies applied to music, and software agents and DAI.
The general architecture of the system we have designed is then described, 
together with the two main modules of the hybrid system.  We also show 
results from experiments performed with each module separately. Finally, we 
demonstrate the influence of both components in experiments run using the 
entire system, NetNeg.

\section{Artificial Intelligence and Music}

In this section we describe some representative research that applies
Artificial Intelligence (AI) methodologies (i.e., symbolic, logic,
connectionist and hybrid approaches) to music.  A detailed overview of
music systems that use AI tools can be found in~\cite{AC93}.

\subsection {The Symbolic Approach} 

In the symbolic approach, knowledge about the world is represented in
an appropriate language.  The symbols described in this language can
be manipulated by a machine to infer new knowledge.

One piece of research that demonstrates the symbolic approach to music
is EMI (Experiments in Music Intelligence) designed by David Cope
(see~\cite{DC91} and a later work~\cite{DC96}).  The EMI project
focuses on the stylistic replication of individual composers.  EMI
discovers patterns shared by two or more compositions from a specific
style. These patterns are called {\em musical signatures}. The
patterns are detected during a pattern ``almost'' matching process (to
distinguish from a regular pattern matching process). Patterns are
weighted by how often they appear, and musical events (e.g., intervals)
are counted and represented as a statistical model.  The program fixes
the signatures in an empty form in the locations where these
signatures were found in the first input work. Then, the program
composes music by filling the spaces between the signatures according
to music rules and the statistical model. Proper interpolation of this
new music relies on an augmented transition network (ATN). The ATN
orders and connects appropriately composed materials according to the
style, and fleshes out a new work.

\subsection {The Logic Approach}

In the logic approach, knowledge is represented using logical
formalisms. New knowledge is inferred by applying logical operations
and manipulating this logic-based knowledge.

A representative piece of research that demonstrates the logic
approach to music is Choral --- an Expert System for Harmonizing
Chorales in the Style of J.S. Bach of Kemal
Ebcioglu~\cite{KE92}. Choral is an expert system that harmonizes
four-part chorales in the style of Bach.

The system contains about 350 rules, written in the form of
first-order predicate calculus (in a logic programming language called
BSL).  The rules represent musical knowledge from multiple viewpoints
of the chorale, such as the chord skeleton and the melodic lines of
the individual parts.  Choral uses an analysis method taken from
Schenker~\cite{Sch69,Sch79} and the Generative Theory of Tonal Music
of Lerdahl and Jackendoff~\cite{LJ83}.

The Choral system harmonizes chorale melodies using the generate and
test method.  The generate section contains condition-action pairs:
all possible assignments to the $n$th element of the partial solution
are sequentially generated via the production rules. The test section
contains constraints --- if a candidate assignment does not comply with
the constraints, it is thrown away. The recommendations section
contains heuristics --- the successful candidates are weighted by the
heuristics and are saved as a sorted list. The program attempts to
continue with the best assignment to the element $n$. If a dead-end is
encountered, a backtracking return is made to this point, with the
next best assignment as defined by the sorted list. The heuristics in
the search are used for biasing the search toward musical solutions
(by changing the order of facts and rules for the search in the data
base and for the backtracking).

The program has reached an acceptable level of competence in its
harmonization capability. Its competence approaches that of a talented
student of music who has studied the Bach chorales. Nevertheless,
problems occur with Schenkerian analysis knowledge base, and a general
criticism is made by experts as to the lack of excitement and global
coherence. The experts also report that they use transformations of
solution prototypes to solve a new problem.

\subsection {The Sub-symbolic Approach}

Connectionist researchers believe that an intelligent machine should
reflect the Central Nervous System (CNS). Knowledge learned by the
machine is expressed by the states and the connections between simple
processing units (neurons).

Artificial Neural Networks (ANN) are information processing systems
(IPS) that have certain performance characteristics in common with
biological neural networks. ANN developed as generalizations of
mathematical models of human cognition or neural biology. The ANN
paradigm enables learning from a set of examples, avoiding the need to
formulate rules. The ANN researcher deals with issues concerning
representation, architectures, and learning algorithms.  Their main
concern is to choose a set of examples that show {\em what\/} is the
requested behavior; they do not need to handle any procedural details
regarding {\em how\/} to solve the problem algorithmically.

ANN assumptions include the following.  Information processing occurs
at many simple elements called neurons. Signals are passed between
neurons through connection links. Each connection link has an
associated weight that typically multiplies the signal
transmitted. The weights represent the information being used by the
net to solve a problem. Each neuron usually applies a nonlinear
activation function to its net input (sum of the weighted input
signals) to determine its output signals.

Listening, performing, and some other musical activities can be
represented using a sequential stream of information. The choice of
Jordan's sequential net (\cite{Jordan86}) is appealing in such cases.
Jordan's sequential net is a version of the back-propagation
algorithm~\cite{RM86}.  Using the learning algorithm, the sequential net
is able to learn and predict sequential elements (such as the sequence
of a melody's notes or harmonic progression).

The sequential net contains three fully-connected layers.  The first
layer contains a pool of state units and plan units.  The second layer
is the hidden layer, and the third layer is the output layer.  The
output layer and the state units contain the same number of units.
The output layer is fed back into the state units of the first layer
for the computation of the next sequential element. The value of a
state unit at time $t$ is the sum of its value at time $t-1$
multiplied by some decay parameter (the value of the decay parameter
is between 0 to 1) and the value of the corresponding output unit at
time $t-1$. The state units represent the context of the current
sequential element, and the output layer represents the prediction of
the net for the next sequential element.

Peter Todd~\cite{PT91} suggested exploiting the Jordan sequential net
for predicting sequential musical elements. His neural network model
presented a connectionist approach for algorithmic composition.  In
the learning phase, the net learns a set of melodies' notes. Each
melody is associated with a unique label encoded in the plan units. In
the generalization phase, new melodies are produced by interpolation
and extrapolation of the labels' values encoded in the plan units. The
resulting melodies share similarities with the melodies within the
learning set. These similarities are unique and different from those
resulting from other methods, and they are interesting from the
compositional aspect.

Two of the authors of this paper together with Naftali Wagner
(see~\cite{GLW98}) suggested the use of a sequential neural network
for harmonizing melodies in real time.  The net learns relations
between important notes of the melody and their harmonies and is able
to produce harmonies for new melodies in real time (i.e., without
knowledge of the continuation of the melody).  The net contains a
sub-net for representing meter that produces a periodic index of meter
necessary for viable interpretations of functional harmonic
implications of melodic pitches.  This neural network model suggests a
method of building a system for automatic generation of real-time
accompaniment in live performance situations. Such a system can be a
basis for enhancing musical electronic instruments, educational
software or other real time interactive applications (such as NetNeg).

\subsection{The Hybrid Approach}

In the hybrid approach, knowledge about the world is represented by an
integration of a sub-symbolic system and a symbolic system (e.g.,
a neural net can be integrated with a rule-based agent system).

A key motivation for the hybrid approach is the assumption that
handling the complexity of AI tasks is beyond the reach of a single
paradigm. Melanie Hilario (see~\cite{Hilario95}) distinguishes among
various hybrid approaches (in her terminology, neurosymbolic
integration).  She suggests classifying approaches into two
strategies: unified strategies and hybrid strategies.

Unified strategies enrich neural networks with symbolic
capabilities. Hybrid strategies combine neural networks and symbolic
approaches at different levels. Hybrid neurosymbolic models can be
either translational or functional hybrids.  Translational hybrid
systems use neural networks as the processors. The symbolic approach
is applied on the network input and targets.  Functional hybrid
systems exploit both the neural network and symbolic components
equally.

The system we present in this work is functional hybrid. Moreover, it
is loosely coupled, since each of the components (i.e., the symbolic
and sub-symbolic) act locally in time and space, and the interaction
between them is always initiated by one of them.  In our case, the
integration of both components is appropriate to the chainprocessing
integration mode as explained in~\cite{Hilario95}. Specifically, we
can look at one of the processes as doing the main task, and the other
as pre/post processing the relevant information. In NetNeg, either of
the two modules can be viewed as the main module, and in charge of the
other.

To the best of the authors' knowledge, the integration of a symbolic
system with a sub-symbolic system, and in particular the integration
of a rule-based agent module with a neural network module for
representing musical knowledge, is novel.

Another known hybrid system is HARP~\cite{camurri95} (Hybrid Action
Representation and Planning). This system has been designed for
computer-assisted composition, performance, and analysis. As explained
in~\cite{camurri95}, HARP is considered hybrid since it combines
different formalisms. The symbolic module consists of a semantic net
together with a temporal logic with production rules. The sub-symbolic
module is composed of a system of cooperative agents. It is not clear
whether the sub-symbolic component in this system handles only the
representation of knowledge, or whether it also handles the processing
and integration of data.  However, NetNeg processes data at a
sub-symbolic level (i.e., the neural network), and also processes data
at the symbolic level (i.e., the rules about which the agents can
reason).  In addition, NetNeg integrates the results of this
processing, namely each component makes use of the results of the
other's as structures for its own input.


\section{Software Agents}

Agents (robots or software agents~\cite{agents95,agents96}) are
functional, independent software modules that are programmed to act on
behalf of the user.  Among the salient features of software agents are
autonomy, adaptation, and sociability.  Agents are autonomous in the
sense that after they are given a goal, they can decide how it would
be achieved in terms of the steps to be taken and the time needed for
its execution.  Agents might act in static as well as in dynamic
environments. An agent might be embedded in a world in which it is the
only software entity, or the agent might need to interact with other
agents of the same or different types (homogeneous or heterogeneous
societies).  In these cases, agents can benefit if they are adaptive;
i.e., agents that learn from their environment and/or the other agents
with which they interact or whose actions interfere.

There has been research on the standardization of communication
languages (e.g., KQML) for agents, allowing agents of differing types
to communicate with one another. Agents might also build beliefs
models of the other agents in the environment and behave
accordingly. Agents have been built that cooperate with one another,
even in cases where communication among them does not exist. This has
been achieved by imposing social laws on their behavior. Agents might
also be capable of communicating with their human users.

Distributed Artificial Intelligence (DAI) is the area in AI that
investigates the behavior of societies of agents. Research has been
divided into two main streams: Cooperative Problem Solving (CPS) and
Multiagent Systems (MAS).\footnote{More recently, ``multiagent
systems'' has been used as another name for the entire field of DAI.}
The main difference between these approaches lies in the global
objective of the system of agents. In CPS systems, agents are assumed
to be cooperative and to have a common goal. The agents have been
designed together, and will assist the others for the group's
benefit. In MAS, agents might have been designed by different
designers.  Agents might be self-interested, and they have their own
personal goals. In these systems, agents might benefit from
cooperating with one another, but their actions or goals might also
conflict.

In heterogeneous environments, even when the agents might be
self-interested and their goals might differ, agents may be able to
coordinate their actions, and also may benefit if they cooperate.

One of the main research areas DAI is concerned with regards applying
coordination protocols to multiagent systems. The agents programmed to
follow these protocols, or agents that are able to learn how to
behave, coordinate with other agents in the same system in order to
achieve global goals, or to avoid conflicts.  To study different
coordination approaches, different models have been studied, including
negotiation~\cite{smith78a,durfee88,kraus91b,zlotkin93b,Rosenschein94},
economics~\cite{malone88,wellman92,sarit93}, social
laws~\cite{Goldman94,tennenholtz89, shoham92a}, and
others~\cite{GenPGP,CLARKE-T,GroszKraus93}.

One way used by agent designers to quantify the agents' performance is
to let the agents compute a utility function. In certain systems, this
function represents the gain of the agent from choosing an action
because of the existence of other agents, in contrast to the cost
incurred by working alone in the world. Other notions of utility might
describe how much information an agent has, or how well the agent does
for a given choice of action.  Whenever agents are self-interested,
they can express their interests and preferences in their utility
function.

In order to achieve the goals assigned to the agents, these agents
choose actions to perform. The algorithms that describe the agents'
behavior take into consideration the utility values, and according to
them, the agents choose their actions. Therefore, the actions that the
agents will perform are strictly related to the computation of the
utility function, which specifies the agents' interests.

The computation of this utility function and its semantics together
with the autonomy, adaptability, and sociability characteristics of
the agents, make a multiagent system richer in expressiveness and in
the possible interactions that might emerge than a static rule-based
system. Agents might change their behaviors to be able to respond
better to other agents operating in the same world. The utility
function might take into consideration facts or terms that would be
difficult to define explicitly by a data base of rules. Therefore, in
this research we implement a utility function approach for our
agents. Later in this article we also show how the function can be
formulated in a different way, so as to express differing interests of
agents (e.g., one agent can be designed to prefer a better melody line
by increasing the weight given to the network recommendation, and
another agent might decrease the weight given to the contrary motion
term).

Research in Distributed Artificial Intelligence (DAI)~\cite{DAIr88}
can also contribute to developing new methods for computer music. For
example, an interesting research question to investigate is the
analogy between the dynamics of the performance of a group of musical
instruments or voices in a vocal ensemble to multiagent systems and
their interactions in the DAI sense.

\section{The Nature of the General Architecture}
\label{architecture}

We distinguish among three aspects that are important for a musician.
The overall design of our system has been guided by these aspects: his
knowledge, intuitions, and aesthetics. The musician's knowledge (e.g.,
rules for a known style, or rules he has invented), intuitions (i.e.,
fuzzy concepts about the music he is interested in composing), and
aesthetic taste (e.g., by learning regularities that appear in the
training examples), can be encoded in different modules.


The system is composed of agents;
we can look at musical activities that can be decomposed into functional
components. All these components interact in order to achieve a
global goal. 
Many musical examples involve such interacting processes: producing
the voice leading of a vocal piece, a string quartet performance, or
composing different parts of a polyphonic melody.
Each such functional component can be implemented and conceptualized
as an autonomous agent. Then, the global goal or activity can be
understood as the goal of the multiagent system and the interactions
among the parts are the social interactions among the agents (e.g., by
coordinating, communicating, teaching).  In our case, the agents
communicate, cooperate and share tasks in order to improve the global
performance of the system. Each agent {\em knows\/} rules of a
specific style, and heuristic rules that take into consideration
different aspects of the problem that the system is trying to
solve. The aesthetic taste referred to above might be captured by a
learning mechanism (e.g., a neural network) that will give advice to
the agents.

An example that we have implemented and will present in the following
sections, refers to the problem of composing polyphonic music in real
time. This implementation demonstrates a specific solution using the
approach presented in this section.

\section{NetNeg's Architecture}

In many musical styles, the composer needs to create different
sequences of notes (i.e., melody lines) that will be played
simultaneously. Each sequence should follow some aesthetic criterion,
and in addition the sequences should sound appropriate when combined.
This overall composition is the result of many interactions among its
components. The musician achieves his overall result by compromising
between the perfection of a single component and the combination of
sequences as a whole. Thus, in this activity there is a constant
tradeoff between the quality of a single sequence versus the quality
of the combined group of sequences.  When a musician is faced with
such a task, he is involved in a cognitive process, that we suggest
might be seen as a negotiation process. He has to compromise between
the melodies' notes by choosing from among the permitted notes those
that are preferable.

\begin{figure}[htbp]
\centerline{\psfig{figure=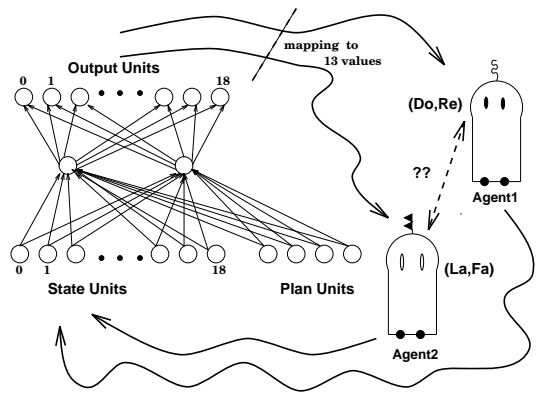,height=2in}}
\caption{The NetNeg Architecture}
\label{system}
\end{figure}

The case study we chose for our experiment deals with first species of
two part counterpoint melodies.
In NetNeg, we create both parts dynamically, in real time. 
Therefore, the system is not allowed to perform backtracking. 
A general view of the architecture of {\em NetNeg\/} is shown in
Figure~\ref{system}.

NetNeg is composed of two sub-systems: a connectionist sub-system 
and a DAI-based subsystem. 
The role of the connectionist subsystem is to learn and generate
individual parts of the polyphonic melody. 
In our implementation, the network learned to reproduce a series of
learning examples that were taken from~\cite{Jeppesen}.
Based on this learning process and the set of learning examples, 
the neural net is able to produce in the generalization phase new 
individual melody parts.
In this phase, the network predicts in the output layer a vector 
of expectations for the next note in each part of the melody. 

Each agent represents one of the voices of the polyphonic music.
It is responsible for choosing the tone that will be inserted
in its voice at each unit of time.
Each agent receives a different output vector from the network.
On the one hand, each agent has to act according to its 
voice's aesthetic criteria; and on the other hand, it has to 
regard the other voice-agent such that both together will result in
a two-part counterpoint. Both agents have to negotiate over 
all the other possible combinations to obtain a globally superior 
result. Thus, they influence the context with their agreement. 
Given this new context and the initial values of the plan units,
the network will predict another output vector.
This process continues sequentially until the melodies are completed.

We describe each of these modules separately and then present results
from experiments performed with NetNeg.

\subsection{The Connectionist Subsystem}
\label{Net}

Each part of the melody is produced independently by a neural network
implemented in Planet~\cite{planet91}. Todd~\cite{PT91} previously
suggested a sequential neural network that can learn and generate a
sequence of melody notes. Currently, our neural network is based on
the same idea, although we have extended it to include the
representation of the contour of the melody.

We built a three-layer sequential net, that learns series of notes.
Each series is a one part melody. Each sequence of notes is labeled by
a vector of plan units. The net is a version of a feedforward
backpropagation net with feedback loops from the output layer to the
state units (in the input layer).  The state units in the input layer
and the units in the output layer represent the pitch and the
contour. The state units represent the context of the melody, which is
composed of the notes produced so far.  The output unit activation
vector represents the distribution of the predictions for the next
note in the melody for the given current context.

The role of the plan units is to label different sequences of
notes. In the generalization phase, we can interpolate and extrapolate
the values of the plan units so as to yield new melodies.  At each
step, the net is fed with the output values of the previous step in
the state units together with the values of the plan units.  These
values will cause the next element in the sequence to appear in the
output layer and it will be propagated as feedback into the state
units (these connections do not appear in Figure~\ref{system}).  The
current values of the state units are composed of the previous values
multiplied by a decay parameter and the current output values.

The state units and the output layer can represent the notes in
different ways.  Each note is represented by a binary vector. The
pitch is associated with the index of the single 1 in the vector
(e.g., RE will be encoded as (10000000) and MI as (01000000) in the
Dorian modus\footnote{The training set as shown in
Figure~\ref{learnEx} is written in Dorian modus. The Dorian modus is
a scale of pitches where the RE is the highest pitch in the pitch
hierarchy. This modus is similar to the pure minor scale and differs
in having a higher sixth pitch.}). In this implementation, we choose to
represent the notes as a vector of 19 units. The first eight units
encode the pitch.  The next nine units represent intervals between the
notes. The last two units describe whether the movement of the melody
is ascendent or descendent (we will refer to these units as the
movement units).

For example, if the current tone is DO(C), the net predicts both RE(D)
and FA(F) as the next best tones, and the ascendent movement unit is
on, then the interval can help us to decide which tone to choose
(i.e., one tone or two and a half tones).  If after the net has chosen
the tone SOL(G), it predicts LA(A) or FA(F) and the interval is of one
tone, then we could choose whether to descend or ascend based on the
activations of the movement units.
In order to exploit the information encoded in the output units' activations,
the pitch activations were combined with the interval and the movement
activations. The activations of the output units were mapped into a
vector of thirteen activations corresponding to the notes in more than
an octave and a half.

Each agent receives the 13-length vector, and feeds the state units
their agreement (see Figure~\ref{system} and
Section~\ref{agents}). Then, the network predicts another output
vector given this new context and the initial values of the plan
units. This process continues sequentially until the melodies are
completed.

\subsection{The DAI-Based Subsystem}
\label{agents}

The agent module was implemented by using the Mice testbed~\cite{mice92}.
In the implementation presented in this work, each voice of the
bicinia is represented by an agent. Since we are dealing with two
counterpoint melodies, then in DAI terms we design a multiagent system
composed of two agents. The global goal of the system is to compose
the two part melody following the rules of the style. In addition,
each single agent has its own individual goal, i.e., to compose its
melody by choosing the {\em right notes}. In particular, each agent has to
act according to the aesthetic criteria that exist for its voice; at
the same time, it has to compose the voice in a manner compatible with
the other voice-agent such that both together will result in a
two-part counterpoint.

At every time unit in our simulations, each agent receives from the
network a vector of activations for all the notes among which it can
choose. Were the agent alone in the system, it would have chosen the
note that got the highest activation from the neural network, meaning
that this note is the one most expected to be next in the melody.
Both agents' choices might conflict with respect to the rules of the
style and their own preferences. Therefore, we apply a negotiation
protocol~\cite{Rosenschein94} to allow the agents to coordinate and
achieve their mutual goals.  The agents will negotiate over all the
other possible combinations to obtain a globally superior result.

In principle, each agent can suggest any of the $n$ possible notes
received from the network. Not all of these pairs of note combinations
are legal according to the rules of the species.  In addition, there
are specific combinations that are preferred over others in the
current context. This idea is expressed in this module by computing a
utility function for each pair of notes. In this sense, the goal of
the agents is to agree on the pair of notes that is legal and also
achieves the maximal utility value among all options.

At each time unit, for each pair of notes, the agents start a
negotiation process at the end of which a new note is added to each of
the current melodies.  Each agent sends to the other all of its notes,
one at a time, and saves the pair consisting of its note and the other
agent's note that a) is legal according to the first species style
rules and b) has yielded the maximal utility so far. At the end of
this process, the pair that has achieved maximal utility is chosen.
Both agents feed their networks with this result as the current
context so that the networks can predict the next output. Each agent,
then, receives a new input based on this output, and the negotiation
step is repeated until the melody is completed.

The term in the utility function that encodes the rules of a given
style expresses (in our implementation) the rules of the polyphonic
vocal style of the sixteenth century as they appeared
in~\cite{Jeppesen}.
A pair of notes is considered legal according to the following rules:
\begin{enumerate}
\cutwspace
\item The intervals between pairs of notes in the two part melodies
should not be dissonant (i.e., the second, fourth, and seventh
intervals are not allowed).
\item There should be perfect consonance (i.e., unison, octave, and perfect 
fifth intervals) in the first and last places of the melody.
\item Unison is only permitted in the first or last places of the melody.
\item Hidden and parallel fifths and octaves are not
permitted.\footnote{Following~\cite{DictMusic}, ``parallel motion of
perfect intervals is forbidden, nor may any perfect interval be
approached by similar motion''.}
\item The difference between the previous and the current interval
(when it is a fifth or an octave) should be two (this is our modification).
\item The interval between both tones cannot be greater than a
tenth.
\item At most four thirds or sixths are allowed.
\item If both parts skip in the same direction, neither of them
will skip more than a fourth.
\item In each part, the new tone is different from the previous
one.
\item No more than two perfect consonants in the two part
counterpoint, not including the first and last notes, are allowed 
(this is our modification).
\end{enumerate}

One example of an aesthetic preference or intuition in our current
implementation is captured by preferring contrary motion.  The
function values will be determined according to whether the pairs of
notes are legal or illegal based on the rules given above, and whether
they are more preferred or less preferred, based on the net advice and
fuzzy concepts given by the musician (e.g., contrary motion).  The
utility function we chose is one example of a function that computes
all the aspects we described in Section~\ref{architecture}.

More formally, we define the utility function as follows:

\label{UtilityFunc}
$SysUtility(T_1^t,T_2^t)=[(act(T_1^t)*act(T_2^t))+cm(T_1^t,T_2^t)]*
match(T_1^t,T_2^t)$

where: $T_i^t,\ i \in \{1,2\}$ is the tone proposed by agent $i$ at time
$t$. $act(T_i^t)$ is the activation of tone $i$ as rated by the neural
network. The term $cm$ characterizes the motion between the previous two
tones and the current pair of tones. We will denote the following
condition $cmCond$ as true when there is contrary motion. Notice that
the $cm$ values are larger whenever the movement steps are smaller.
$cmCond(T_1^t,T_2^t) \stackrel{\rm def}{=}\neg(((T_1^t < T_1^{t-1})
\wedge (T_2^t < T_2^{t-1}))\vee ((T_1^t > T_1^{t-1}) \wedge (T_2^t >
T_2^{t-1})))$

Then we define
$cm(T_1^t,T_2^t)=\left\{ \begin{array}{lll}
	\frac{1}{|interval(T_1^{t-1},T_2^{t-1}) - interval(T_1^t,T_2^t)|} 
&\mbox{ if $cmCond(T_1^t,T_2^t)$ }\\
                      0 &\mbox{otherwise}
                     \end{array}
            \right. $

$match(T_1^t,T_2^t)=\left\{ \begin{array}{ll}
				1 &\mbox {if $(T_1^t,T_2^t)$ is 
legal w.r.t.~the above rules}\\
				0 &\mbox{otherwise}
			    \end{array}
		     \right. $

We have considered contrary motion in this function because this type
of motion produces the most natural and appropriate effect for this
kind of music (as noticed by Jeppesen~\cite{Jeppesen}).  An
interesting point to notice is that taking into account the contrary
motion term in the function enables the system to produce contrary
motion although the network might have suggested movement in the same
direction (based on the cantus firmi with which the network was
trained).

In this implementation, both agents' neural networks were trained with
the same set of training examples, and both compute the same utility
function.  In a more general case, we might have a more complex, and
richer, multiagent system by implementing agents with different utility
functions (i.e., the agents combine or weight in different ways the same
terms, or consider other terms to compute the aspects discussed in
Section~\ref{architecture}).  In this way, agents might stick to a
specific kind of behavior; agents with different {\em character\/}
might be implemented (e.g., an agent that wants its voice to be more
salient along the melody).

One way to enrich the kinds of melodies produced is to enable
nondeterministic rules. This can be expressed by having a system where
one agent can compute the utility of the same pair of notes for the
system in different ways. One example is to have a system composed of one
agent that prefers more contrary motion than the other. In our case
this will be expressed by giving more weight to the $cm$ term in the
computation of the utility function.  
This weight might be chosen from a certain distribution.
Another example is to have agents that prefer X\% of the time skips, and the
other prefers Y\% steps.

More generally, we can think of each agent as having two different
functions. In other words, each agent sees the utility function as a
computation of different terms that should be considered.
This is an example of a multiagent system in which each agent
could be designed by a different designer, and has its own interests
and desires. The benefit of the whole system is not necessarily
the benefit of each of the agents. 

\section{Experiments}
   
We first ran each subsystem separately to examine the ability of each
one of the two approaches (i.e, Neural Nets and agents) to cope with
the general problem. We then ran the integrated system; we present
results from all of these simulations.  In this way, we show the
ability of the whole system to produce results superior to the
performance of either of the subsystems. Combining the modules gave us
a more natural way of dealing with the processing of and
representation of our task.

\subsection{Running the Net Module}

The task of the net was to learn to produce new two-part melodies.
This case is different from the one faced by the whole system, 
in which only one-part melodies were taught.
Therefore, we needed to represent both parts of the melody 
simultaneously.
We used the same sequential net that was described in 
Subsection~\ref{Net}.
In this case, we doubled the number of the units in each layer to 
represent two notes simultaneously, one for each part. 
In the learning phase the net was given four melodies, containing 
the two parts. One example from this set of melodies follows:
\newline
{\tt V1:$re8\  do8\  mi8\ la\ do8\ si\ la\ do8\ re8$}\newline
{\tt V2:$re8\ la8\ sol8\ fa8\ mi8\ re8\ fa8\ mi8\ re8$}\newline

Since our notes are taken from one and a half octaves, we represent
the notes by their names (i.e., re), and those in the higher octave
have an 8 concatenated to their names (i.e., re8).
In this phase, the net learned the examples in the set with 
high accuracy after short training.\footnote{For 20 hidden 
units it took less than 100 epochs to achieve an average error 
around $0.0001$.}
Each melody had a different label encoded as a unique value in the 
plan units. After training, we tested the net by supplying as input
the four labels 1000, 0100, 0010, 0001, one for each of the four 
melodies in the set, and the net was able to completely reproduce
the sequence without mistakes.

In the generalization phase we chose to interpolate the 
values of the plan units to produce new melodies. Todd~\cite{PT91}
demonstrated that the resulting sequences have non-linear 
similarities to the sequences in the learning set depending on the
activations of the plan units. 
An example of a typical result follows:\newline
The plan vector: (0.3 0.7 0.3 0.7). \newline
{\tt V1:$re8\ sol\ si\ do8^{**}\ mi\ re^{***}\ do8\ re8$}\newline
{\tt V2:$re8\ fa8^{*}\ sol8\ la8\ sol8\ fa8\ mi8\ re8$}


This resulting sequence reflects typical problems we encountered
when dealing with this simple approach. The examples in the 
learning set imposed two different constraints on the net.
The constraints regard the melodic intervals between the pitches
in each part, and the combinations of pitches in both parts.
The net is not able to cope with both constraints consistently, and
thus it satisfies each, one at a time. 
For example in $*$, the combination chosen is not allowed in the
specific style, although the melodic interval is fine.
In $**$ the descending skip is not permitted, as well as the
ascending skip in $***$, but both combinations are fine. 

\subsection{Running the Agent Module}

The agents in our system know the rules of the specific style of the
melodies we want to compose. They also know how to compute the system
utility for a given pair of notes.  We have run experiments with the
agent module alone. We remove the influence of the recommendations
produced by the neural network by giving the agents a vector of zero
activations for all the possible notes. In this way, we wanted to
check that the voices we will get by solely applying knowledge about
interactions between the two parts will lack the features learned
by the net in its training phase (i.e., the aesthetics of one part).

We run the module with the utility function described in 
Section~\ref{UtilityFunc}, where the net's advice was assigned
zero.
Since we choose the pair of notes that get the maximal
utility value at each step, the result is:\newline
{\tt V1:$re8\ do8\ la\  sol\  la\  mi\   la\ re8$} \newline
{\tt V2:$re8\ mi8\ fa8\ sol8\ fa8\ sol8\ fa8\ re8$}\newline

The melody lacks the features requested from each part.  In both
voices there are redundant notes (i.e, the appearance of note la in
V1, and the series of notes in V2 from the third place to the
seventh). There is no unique climax in any of the voices.  There are
two continuous skips in the last three notes in V1.  There are too
many steps in V2 (i.e., there is no balance between the skips and the
steps).

We also observed that melodies can come to a dead end, when there is no
pair of notes that can satisfy the rules of the specific style.  In
such cases, we have tested additional results that can be produced
by the system when the agents are allowed to choose pairs with utility
values smaller than the maximum.


\subsection{Running NetNeg}

\begin{figure}[htbp]
\centerline{\psfig{figure=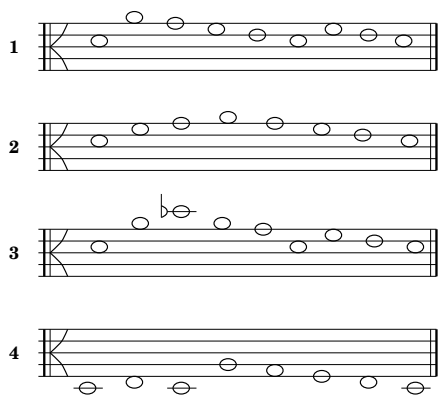,height=2in}}
\caption{The learning examples}
\label{learnEx}
\end{figure}

In this section we present the main simulation performed on the whole
system.  In the training phase, the network learned to reproduce four
melodies that were taken from~\cite{Jeppesen}. See Figure~\ref{learnEx}.

We have tested the performance of the network with different learning
parameters, such as the number of hidden units and the values of the
plan units.  The results we present were produced by a net with 15
units in the hidden layer.

In the generalization phase, given a specific vector of plan units,
the network produces a new cantus firmi. We have chosen two different
plan vectors for the net that will output the notes for each agent. We
run the net, each time with the corresponding plan vectors, and mapped
their outputs to two different thirteen activation values. Then, we
run the DAI-based module with these inputs. The agents negotiate over
the different pairs of possible combinations, computing for each the
system utility.  Finally, the agents agree upon a legal pair of notes
that has yielded the maximal utility; alternatively, the agents might
decide that no combination is legal, given the previous note in the
melody.

In our current case, the nets are fed with the agents' agreement
and the system continues to run. This process is executed until 
the two-part melodies are completed. Currently, the length of the
melodies is fixed.

\begin{figure}[htbp]
\centerline{\psfig{figure=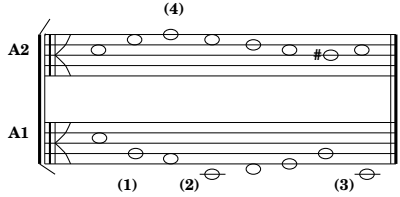,height=1in}}
\caption{A new two-part melody}
\label{result}
\end{figure}

A melody that resulted from an experiment we performed is shown
in Figure~\ref{result}.
The net was presented with two different plan vectors 
((0.8 0 0.8 0) and (0 1 0 1)). The agents computed the utility of
the system taking into account the rules described
in Section~\ref{agents} and the contrary motion term.

In Figure~\ref{result} we can observe that the system gives aesthetic
results, quite appropriate for the species style with which we have
experimented.  Both parts are consistent with the combination
constraint, as opposed to the simulation we ran solely with the neural
network, where this constraint was not satisfied. Comparing with the
simulation run with the agents alone, no redundancy was found in this
example.  Nevertheless, there is a contour problem as pointed out in
(1) and (2) in A1's melody in Figure~\ref{result}.  According to
Jeppesen~\cite{Jeppesen}, it is preferred to descend by a step and
then perform a descending skip. After a descending skip, we are
expected to have a compensating ascending movement.  In (3), we prefer
to approach the last note by a step. A2's melody is perfect with
regard to the contour. There is a single climax as shown in (4).

By running both modules together, we see the combined effect of all the 
aspects taken into consideration in the utility function.
The melody of agent $A_2$ is perfect. This reflects the melodies
learned by the neural net. The contrary motion term expressed in our
heuristic is revealed in the relation between both melodies. The
neural network was not trained with melodies such as the one created by
agent $A_1$. The rules are expressed, for example, in the consonance of
the parts of the melody.

The whole system was run only for a deterministic case. The
nondeterministic case was run only to show that NetNeg is also capable
of having agents computing different functions.
%
%
In the nondeterministic case, the agents weighted the contrary
motion term according to a coin toss between $0.5$ and $1.49$.
We did three simulations, and got the following melodies:
\begin{enumerate}
\item {\tt V1:$re8 \ mi8 \  sol8 \ la8 \  sol8 \  mi8 \ fa8 \ re8$} \newline
      {\tt V2:$re8 \ do8 \ si \  la \  si \  do8 \ la \  re8$}
\item {\tt V1:$re8 \ do8 \ la \ sol \ la \  do8 \ si \  re8$}\newline
      {\tt V2:$re8 \ mi8 \ fa8 \ sol8 \ fa8 \ mi8 \ sol8 \ re8$} 
\item {\tt V1:$re8 \ mi8 \ sol8 \ la8 \ sol8 \ fa8 \ sol8 \ re8$}\newline
      {\tt V2:$re8 \ do8 \ si \  la \  si \  re8 \ si \  re8$}
\end{enumerate}

In all the examples produced, there are too many steps in the second
voice. There is no unique climax in the second voice as well.  In
general, the first voices are slightly more interesting.  In the first
example, the second voice is not very interesting; it has too many
steps. There is not any balance among the steps and the skips. The
only skip is at the last note, exactly where we prefer it less.  In
the second example there is not any climax in any of the voices.  In
the third example there is a redundant series in the second voice, (si
re8 si re8).  The analysis of the interactions among specific types of
agents remains for future research.

%
%

\section{Summary and Future Work}

The main contribution of this research is in presenting a powerful
computational approach for representing musical knowledge.

We have presented a novel computational approach for encoding the
knowledge, intuitions, and aesthetic taste of a musician in different
modules.  In this work, we presented an example of an implementation
that enables a human to flexibly guide the system to compose music in
a style he chooses, under real-time constraints.  The user might
express multiple views and levels of knowledge to this system.  For
example, if he knows examples of the music he wants the system to
compose, he can train the neural network with these examples. If he
wants the music to follow specific rules he might formulate them in
the agent module.  In addition, he can regard other factors in the
computation of the utility function.

The system is composed of a connectionist module and an agent-based
module. The agents decide upon the notes in both melodies. The neural
network predicts the expectation for every note to be the next one in
the melody. This vector of expectations is passed as input to each of
the agents.  Each agent knows the rules of the style, heuristics, and
the net's advice.  Based on these, the agents then negotiate over the
possible combinations of notes until they reach an agreement on notes
that, added to the melody, will most greatly ``benefit'' the
system. The pair of notes that has been agreed upon is sent back to
the neural networks to build their new context.  As a case study, we
built a system for composing two-part counterpoint melodies.

We have implemented a specific utility function, but nevertheless our
architecture is general enough to run different kinds of functions for
achieving other tasks.  In a multiagent system, agents may be
self-interested. This can be expressed by giving each agent a different
utility function or nondeterministic functions. We refer to a function
as nondeterministic when, for the same pair of notes, it will return
different values. In our work, we started to investigate functions in
which the contrary motion term was weighted by a coin toss.

Issues to be further investigated include other ways to integrate both
modules, the study of other species (second, third, and fourth
species), and polyphonic music in more flexible and more abstract
styles.  We are also interested in examining how other representations
and coordination protocols~\cite{Goldman95a,Goldman97e,Goldman98d} can
influence the performance of the system.


\section*{Acknowledgments}
We would like to thank Gil Broza for implementing a C++ version of NetNeg.


\bibliographystyle{alpha}

\newcommand{\etalchar}[1]{$^{#1}$}

\end{document}